\documentclass[letterpaper, 10 pt, conference]{ieeeconf}
\usepackage[latin9]{inputenc}
\usepackage{float}
\usepackage{amsmath}
\usepackage{graphicx}

\makeatletter

\DeclareTextSymbolDefault{\textquotedbl}{T1}
\floatstyle{ruled}
\newfloat{algorithm}{tbp}{loa}
\providecommand{\algorithmname}{Algorithm}
\floatname{algorithm}{\protect\algorithmname}

\IEEEoverridecommandlockouts                              % This command is only needed if
\author
{
Liang Sun$^{1}$ and Leonardo Escamilla$^{1}$
\thanks{$^{1}$Liang Sun and Leonardo Escamilla are with Department of Mechanical and Aerospace Engineering, New Mexico State University, Las Cruces, NM 88003, {\tt\small lsun@nmsu.edu}; {\tt\small brwnl30@nmsu.edu}}
}
\usepackage{setspace}
\usepackage{graphics} % for pdf, bitmapped graphics files
\usepackage{epsfig} % for postscript graphics files
\usepackage{mathptmx} % assumes new font selection scheme installed
\usepackage{times} % assumes new font selection scheme installed
\usepackage{amsmath} % assumes amsmath package installed
\usepackage{amssymb}  % assumes amsmath package installed
\usepackage{epstopdf}
\usepackage{cite}
\usepackage[final]{changes}
\usepackage{xcolor}
\usepackage{bbm}
\usepackage{tikz}
\usetikzlibrary{arrows}
\usepackage{caption}
\usepackage{algpseudocode}
\usepackage{mathtools}
\usepackage{algorithm}
\usepackage[para]{threeparttable}
\usepackage{color}

\usepackage{mathtools, cuted}

\makeatother

\begin{document}
\title{\LARGE\bf{Uncertainty-Aware Task Allocation for Distributed Autonomous Robots}}

\maketitle
\thispagestyle{plain} 
\pagestyle{plain}

\begin{abstract}This paper addresses task-allocation problems with
uncertainty in situational awareness for distributed autonomous robots
(DARs). The uncertainty propagation over a task-allocation process
is done by using the Unscented transform that uses the Sigma-Point
sampling mechanism. It has great potential to be employed for generic
task-allocation schemes, in the sense that there is no need to modify
an existing task-allocation method that has been developed without
considering the uncertainty in the situational awareness. The proposed
framework was tested in a simulated environment where the decision-maker
needs to determine an optimal allocation of multiple locations assigned
to multiple mobile flying robots whose locations come as random variables
of known mean and covariance. The simulation result shows that the
proposed stochastic task allocation approach generates an assignment
with 30\% less overall cost than the one without considering the uncertainty.

\end{abstract}

\section{Introduction}

One of the great challenges facing modern society is to develop new
technologies that transform how we manage the transportation ecosystem
to embrace upcoming large-scale distributed autonomous robots~(DARs),
such as unmanned aerial vehicles (or drones) and self-driving cars.
Autonomous robots have demonstrated their capabilities to complete
missions that are too dangerous, dumb, costly, or impossible for humans
to handle, such as transportation~\cite{Hackenberg2020,waymo}, warehouse
sorting~\cite{warehouse_robots}, emergency response~\cite{emergency_drone},
and hazardous chemical spraying~\cite{spraying_drone}, to name a
few. These challenging missions usually consist of multiple tasks
with demands and uncertainty that may vary over time, which pose critical
challenges to the coordination of DARs. The advances of optimization
theory and computer algorithms have revolutionized approaches (e.g.,
market-based algorithms~\cite{Bertsekas1990} and bipartite matching
algorithms~\cite{Kuhn1955}) to centralized task-allocation strategies
for autonomous robots, but these approaches are not directly applicable
to DARs. When distributed optimization techniques are paired with
probability theory and appropriate coordination mechanisms, potentially
transformative gains in stochastic task-allocation and dynamic reallocation
for DARs are possible. However, the widespread use of DARs for real-world
missions has remained elusive, in large part due to limited system-level
understanding and control of the complex changes that DARs undergo
dynamic and uncertain environments.

\begin{figure}[h]
\centering\includegraphics[width=1\linewidth]{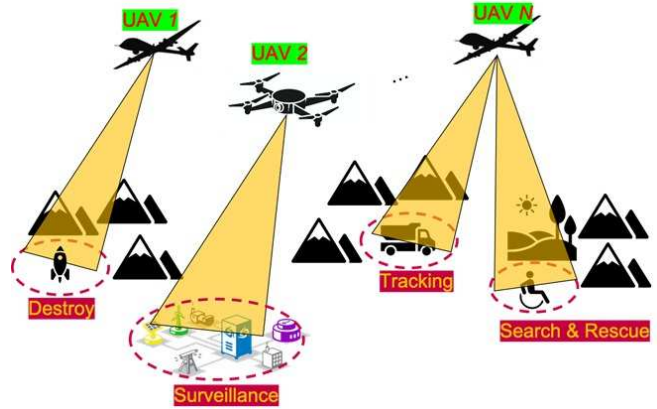}
\caption{A motivating scenario where a team of (heterogeneous) unmanned aerial
vehicles (UAVs) collaboratively executes multiple ``tasks'', which
can be defined as simple actions (e.g., take pictures) or complicated
missions (e.g., mobile-target tracking, search and rescue, etc.) that
may contain subtasks. These tasks and subtasks may be located at distant
places with different priorities and different costs to execute. The
yellow triangles denote the sensing range of the UAVs and sensor measurements
are assumed to be with uncertainty. }
\label{droneImage}
\end{figure}

Figure~\ref{droneImage} depicts a motivating scenario for this paper,
where a team of (heterogeneous) DARs (unmanned aerial vehicles (UAVs)
in this case) collaboratively executes multiple ``tasks'', which
can be defined as simple actions (e.g., take pictures) or complicated
missions (e.g., mobile-target tracking, search and rescue, etc.) that
may contain susituaionalbtasks. These tasks and subtasks may be located
at distant places with different priorities and different costs to
execute. In real-world scenarios, the state of DARs and tasks are
full of uncertainty, and it is always preferred by a (human) decision-maker
to identify the successful rate of executing a series of tasks. 

As an integral part of combinatorial optimization~\cite{PapadimitriouSteiglitz1998,MartelloMinouxRibeiroEtAl2011},
task allocation processes often form building blocks of solutions
to complex problems and have been widely investigated in the literature~\cite{JiAzumaEgerstedt2006,GerkeyMataric2004,SmithBullo2007,BurkardDell'AmicoMartello2009},
such as resource allocation and coordination of DARs~\cite{TurraPolliniInnocenti2004,JinLi2018,MayyaWilsonEgerstedt2019,McLurkinYamins2005}.
Auction and Hungarian algorithms are two well-known centralized task-allocation
methods in their original forms, which have been initially applied
in assignment problems. The Hungarian algorithm~\cite{Kuhn1955}
was the first to compute an optimal solution in finite time to the
linear sum assignment problem (LSAP)~\cite{BurkardCela1999} and
has been implemented in~\cite{papadimitriou1998combinatorial,munkres1957algorithms}.
It has been utilized in applications, such as moving-target detection
and tracking~\cite{Tanner2007,TurraPolliniInnocenti2004,BewleyGeOttEtAl2016},
formation generation~\cite{JiAzumaEgerstedt2006}, and fault tolerance
for cooperative unmanned systems~\cite{KamelGhamryZhang2015}. The
market-based auction algorithm depends on a central auctioneer that
is responsible for assigning tasks according to received bids~\cite{Bertsekas1990,JiAzumaEgerstedt2006,GiordaniLujakMartinelli2013}.
Auction algorithms require robots to bid on tasks, with rewards based
on corresponding prices~\cite{Bertsekas1990,GerkeyMataric2002}.
The Auction algorithm has been applied in various applications~\cite{ChoiBrunetHow2009,DeVriesVohra2003,KangParkes2007,LagoudakisBerhaultKoenigEtAl2004,Milgrom2000,SarielBalch2005,ZavlanosSpesivtsevPappas2008,ZhengKoenigTovey2006},
such as multi-robot coordination~\cite{Gurfil2005,BerertonGordonThrun2004}
and the network flow problems~\cite{BertsekasHoseinTseng1987,Bertsekas1990}. 

Centralized task-allocation algorithms rely on a central agent to
achieve an optimal assignment by considering the situational awareness
of all agents. These algorithms are usually able to produce global
optimal solutions, but unfortunately they have limitations in robustness
and scalability. Distributed auction-based approaches have been widely
explored, such as the Consensus-based bundle algorithm (CBBA)~\cite{ChoiBrunetHow2009},
which has been adopted as a benchmark technique for multi-task allocation
problems. Consensus-based approaches~\cite{Brunet2008,ChoiBrunetHow2009,Hunt2014,SmithWetherallWoodheadEtAl2014,Tahbaz-SalehiJadbabaie2006,Buckman2018}
typically require the robots to converge on a consistent situational
awareness before performing the assignment. Though such methods are
robust, they are typically slow to converge and require the transmission
of large amounts of data. Although the original Hungarian method outperforms
the auction method on the basis of scalability and system requirements~\cite{NarayananNagarathnamMeyyappanEtAl2000},
the distributed version of the Hungarian method has drawn limited
attention~\cite{ChopraNotarstefanoRiceEtAl2017,IsmailSun2017,SamieiIsmailSun2019,SamieiSun2020,LindsaySun2020,lindsay2021sequential}.
Also, most existing algorithms for distributed task allocation assume
that the convergence speed of the algorithm is faster than the change
in system states (e.g., locations of mobile DARs) nor take system
uncertainty into the optimization process for task allocation.

The ubiquitous uncertainty in situational awareness makes deterministic
models difficult to meet the requirement for evaluation of success
for critical missions, thereby stochastic models have been widely
adopted in DAR autonomy. The uncertainty in situational awareness
is usually represented by random variables with specific probability
distributions, e.g., the mean-variance pair, $\left(\mu,\sigma^{2}\right)$.
The propagation of a random variable through a nonlinear function
and the fusion of two random variables have been studied and utilized
in recursive Bayesian filtering techniques~\cite{KokkalaSolinSaerkkae2015},
which were originally developed for probabilistic inference .

Probabilistic inference is the problem of estimating the hidden variables
(states or parameters) of a system optimally and consistently when
a set of noisy or incomplete observations of the system becomes available~\cite{Wan2006}.
The optimal solution to this problem is given by the recursive Bayesian
estimation algorithm, which recursively updates the posterior density
of the system state as new observations arrive. This posterior density
constitutes the complete solution to the probabilistic inference problem
and allows us to calculate any ``optimal'' estimate of the state.
Unfortunately, for most real-world problems, the optimal Bayesian
recursion is intractable and approximate solutions must be used. Within
the space of approximate solutions, the extended Kalman filter (EKF)
has become one of the most widely used algorithms with applications
in state and parameter estimation~\cite{Al-RadaidehSun2019,GalaLindsaySun2018,GalaLindsaySun2018a,GalaSun2019,SunPack2016a}.
Unfortunately, the EKF is based on a sub-optimal implementation of
the recursive Bayesian estimation framework applied to Gaussian random
variables. This can seriously impact the accuracy or even lead to
divergence of any inference system that is based on the EKF.

Over the past two decades, several novel, more accurate, and theoretically
better motivated algorithmic alternatives to the EKF have surfaced
in the literature. These include the Unscented Kalman Filter (UKF)~\cite{JulierUhlmannDurrant-Whyte1995}
and Central-Difference Kalman Filter (CDKF)~\cite{NoRgaardPoulsenRavn2000},
which are collectively referred to as Sigma-Point Kalman Filters (SPKF).
Unlike the EKF that uses a first-order ``linearization'' approach
to deal with nonlinear systems, the Sigma-Point methods involve deterministic
sampling-based approximations of the relevant Gaussian statistics
to achieve \textquotedbl second-order\textquotedbl{} or higher accuracy.
Remarkably, the computational complexity of an SPKF is the same order
as that of the EKF. Furthermore, implementing an SPKF is often substantially
easier and requires no analytic derivation of Jacobians (gradients)
as in the EKF.

In this paper, we proposed a novel solution framework for generic
task allocation with uncertainty for DARs. The proposed framework
decouples the uncertainty propagation and the task allocation processes
by utilizing the sigma-point sampling. It has great potential to be
employed for generic task-allocation schemes, in the sense that there
is no need to modify an existing task-allocation method that has been
developed without considering the uncertainty in the situational awareness.
The proposed framework was tested in a simulated environment where
the decision-maker needs to determine an optimal allocation of multiple
locations assigned to multiple mobile flying robots whose locations
come as random variables of known mean and covariance. The simulation
result shows that the proposed stochastic task allocation approach
generates an assignment with 30\% less overall cost than the one without
considering the uncertainty.

The rest of the paper is organized as follows. In Section~\ref{sec:Problem-Formulation},
the formal formulation for the proposed work is provided. In Section~\ref{sec:Methodology},
we present the methodology of the proposed framework. In Section~\ref{sec:Simulation-Results},
we present the simulation results that validate the effectiveness
of the proposed method and the concluding remarks are summarized in
Section~\ref{sec:Conclusion}.

\section{Problem Formulation\label{sec:Problem-Formulation}}

Considering a set of $m$ DARs, a stochastic representation of the
time-varying situational awareness at time step $t_{k}$ is formally
defined by $\mathbb{S}_{k}\left(\Xi,\mathbf{P}_{\Xi},\mathcal{G}_{c},\mathbb{N}\right)$,
where
\begin{itemize}
\item $\Xi=\left\{ \xi_{1},\xi_{2},\cdots,\xi_{m}\right\} \in\mathbb{R}^{n\times m}$
is a set of $m$ random variables (column vectors), $\xi_{i}\in\mathbb{R}^{n}$,
$i=1,2,\cdots,m$, representing the situational awareness that each
robot possesses, which includes robot states (e.g., position, velocity,
orientation, etc.), states of objects of interest (e.g., a missing
person that is being searched for, a moving object that is being monitored,
etc.), and environmental states (e.g., temperature, humidity, wind
speed, etc.).
\item $\mathbf{P}_{\Xi}=\left\{ \mathbf{P}_{\xi_{1}},\mathbf{P}_{\xi_{2}},\cdots,\mathbf{P}_{\xi_{m}}\right\} $
is a set of covariance matrices, $\mathbf{P}_{\xi_{i}}\in\mathbb{R}^{n\times n}$,
associated with set $\Xi$.
\item $\mathcal{G}_{c}=\left[g_{ij}\right]\in\mathbb{R}^{m\times m}$ is
an adjacency matrix denoting the communication topology among DARs.
\item $\mathcal{\mathbb{N}}=\left\{ N_{1},N_{2},\cdots,N_{m}\right\} $
is a set of neighbor lists for each robot and $a_{j}\in N_{i}$ if
$g_{ij}=1$.
\end{itemize}
Note that the time variable $t$ is omitted in $\left\{ \Xi,\mathbf{P}_{\Xi},\mathcal{G}_{c},\mathcal{\mathbb{N}}\right\} $
for simplification.

For a non-fully connected DAR system (i.e., a robot not directly connected
to all other robots), it takes some time (e.g., several hops) for
updated knowledge to be transmitted from robot $a_{i}$ to another
robot $a_{j}$, where $a_{j}\notin N_{i}$, via the communication
network defined by $\mathcal{G}_{c}$. So the situational awareness
that robot $a_{i}$ possesses about the entire DAR system is formally
defined by $\mathbb{S}_{k}^{i}\left(\Xi^{i},\mathbf{P}_{\Xi^{i}},\mathcal{G}_{c},\mathbb{N}\right)$,
$i\in\mathcal{I}$, where $\mathcal{I}=\left\{ 1,2,\cdots,m\right\} $.
In such an asynchronous situation, the situational awareness among
robots is not necessarily the same at each time step. So let us assume
that $\mathbb{S}_{k}^{i}\neq\mathbb{S}_{k}^{j}$ until the DAR system
converges to a common $\mathbb{S}$. Note that, for simplicity of
presentation of the proposed framework, $\mathcal{G}_{c}$ and $\mathcal{N}$
are assumed static. 

\subsection{Uncertainty Propagation}

\emph{Unscented transform}~\cite{VanZandt2001,Julier2002,KokkalaSolinSaerkkae2015,SunPack2016a}
is a mathematical function used to estimate the result of applying
a given nonlinear transformation to a probability distribution that
is characterized only in terms of a finite set of statistics. The
most common use of the unscented transform is in the nonlinear projection
of mean and covariance estimates in the context of nonlinear extensions
of the Kalman filter.

Define the dynamics of a system using a state-space model as 
\begin{equation}
\mathbf{x}_{k+1}=f\left(\mathbf{x}_{k}\right)+\mathbf{w}_{k},
\end{equation}
and a generalized nonlinear function of states as 
\begin{equation}
\mathbf{y}_{k}=h\left(\mathbf{x}_{k}\right)+\mathbf{v}_{k},
\end{equation}
where $\mathbf{x}_{k}\in\mathbb{R}^{L_{\mathbf{x}}}$is the state
vector, $\mathbf{y}_{k}\in\mathbb{R}^{L_{\mathbf{y}}}$is the output
vector, and $\mathbf{w}_{k}\in\mathbb{R}^{L_{\mathbf{x}}}$ and $\mathbf{v}_{k}\in\mathbb{R}^{L_{\mathbf{y}}}$
are independent, zero-mean Gaussian noise processes of covariance
matrices $\mathbf{P}_{w}$ and $\mathbf{P}_{v}$, respectively. An
optimal estimate of $\mathbf{x}_{k}$ is given by 
\begin{equation}
\hat{\mathbf{x}}_{k}=\mathbb{E}\left[\mathbf{x}_{k}\right],
\end{equation}
where $\mathbb{E}\left(\cdot\right)$ denotes the expected value operation
and the error covariance matrix is given by
\begin{equation}
\mathbf{P}_{\tilde{\mathbf{x}}_{k}}\triangleq\mathbb{E}\left[\left(\mathbf{x}_{k}-\hat{\mathbf{x}}_{k}\right)\left(\mathbf{x}_{k}-\hat{\mathbf{x}}_{k}\right)^{T}\right],
\end{equation}
Assuming a Gaussian distribution of $\mathbf{x}_{k}$, an optimal
estimate of $h\left(\mathbf{x}_{k}\right)$ is given by 
\begin{equation}
\mathbb{E}\left[h\left(\mathbf{x}_{k}\right)\right]=\int h\left(\mathbf{x}_{k}\right)\mathcal{N}\left(\mathbf{x}_{k}|\hat{\mathbf{x}}_{k},P_{\tilde{\mathbf{x}}_{k}}\right)d\mathbf{x},
\end{equation}
where $\mathcal{N}\left(\mathbf{x}_{k}|\hat{\mathbf{x}}_{k},\mathbf{P}_{\tilde{\mathbf{x}}_{k}}\right)$
is a multi-dimensional Gaussian density with mean $\hat{\mathbf{x}}_{k}$
and covariance matrix $P_{\tilde{\mathbf{x}}_{k}}$. The unscented
transform approximates $\mathbb{E}\left[h\left(\mathbf{x}_{k}\right)\right]$
using multi-dimensional generalizations of Gaussian quadratures, also
referred to as Gaussian cubatures, i.e., $\mathbb{E}\left[h\left(\mathbf{x}_{k}\right)\right]\approx\underset{i}{\sum}\omega_{k,i}h\left(\mathcal{X}_{k,i}\right)$,
where weights $\omega_{k,i}$ and sigma points $\mathcal{X}_{k,i}$
are functions of mean $\hat{\mathbf{x}}_{k}$ and covariance matrix
$\mathbf{P}_{\tilde{\mathbf{x}}_{k}}$. Formally, the set of sigma
points, a $(2L_{x}+1)$-tuple located in the center and on the surface
of an $n$-sphere, is defined by 
\begin{align}
\mathcal{X}_{k} & =\left\{ \mathcal{X}_{k,0},\mathcal{X}_{k,1},\cdots,\mathcal{X}_{k,2L_{x}}\right\} \nonumber \\
 & =\left\{ \hat{\mathbf{x}}_{k},\hat{\mathbf{x}}_{k}+\gamma\sqrt{P_{\tilde{\mathbf{x}}_{k}}},\hat{\mathbf{x}}_{k}-\gamma\sqrt{P_{\tilde{\mathbf{x}}_{k}}}\right\} ,
\end{align}
where $\gamma$ is a scalar that determines the spread of the sigma-points
around $\hat{\mathbf{x}}_{k}$ and $\Sigma_{\tilde{\mathbf{x}}_{k}}$~\cite{VanDerMerweWanJulier2004}.
Then the evolution of $\hat{\mathbf{x}}$ and its covariance matrix
is given by 
\begin{align}
\mathcal{X}_{k+1,i} & =f\left(\mathcal{X}_{k,i}\right),\label{eq:sigma_f}\\
\hat{\mathbf{x}}_{k+1} & =\sum_{i=0}^{2L}\omega_{i}^{m}\mathcal{X}_{k+1,i},\\
\mathbf{P}_{\tilde{\mathbf{x}}_{k+1}} & =\sum_{i=0}^{2L}\sum_{j=0}^{2L}\omega_{ij}^{c}\mathcal{X}_{k+1,i}\mathcal{X}_{k+1,j}^{T},\\
i & =0,\cdots,2L,
\end{align}
and the evolution of $\hat{\mathbf{y}}$, its covariance matrix, and
the cross-correlation matrix of $\mathbf{x}$ and $\mathbf{y}$ are
given by
\begin{align}
\mathcal{Y}_{k,i} & =h\left(\mathcal{X}_{k,i}\right),\label{eq:sigma_h}\\
\hat{\mathbf{y}}_{k} & =\sum_{i=0}^{2L}\omega_{i}^{m}\mathcal{Y}_{k,i},\\
\mathbf{P}_{\mathbf{y}_{k}} & =\sum_{i=0}^{2L}\sum_{j=0}^{2L}\omega_{ij}^{c}\mathcal{Y}_{k,i}\mathcal{Y}_{k,j}^{T},\\
\mathbf{P}_{\mathbf{x}_{k}\mathbf{y}_{k}} & =\sum_{i=0}^{2L}\sum_{j=0}^{2L}\omega_{ij}^{c}\mathcal{X}_{k,i}\mathcal{Y}_{k,j}^{T},
\end{align}
where $\omega_{i}^{m}$ and $\omega_{ij}^{c}$ are scalar weights.

The \emph{sigma-point approach} differs substantially from general
stochastic sampling techniques, such as the Monte-Carlo integration,
which requires orders of magnitude more sample points in an attempt
to propagate an accurate (possibly non-Gaussian) distribution of the
state. The deceptively simple sigma-point approach results in posterior
approximations that are accurate to the third order for Gaussian inputs
for all nonlinearities. For non-Gaussian inputs, approximations are
accurate to at least the second-order, with the accuracy of third
and higher-order moments determined by the specific choice of weights
and scaling factors. Furthermore, no analytical Jacobians of the system
equations need to be calculated as is the case for the EKF. This makes
the sigma-point approach very attractive for use in \textquotedblleft black
box\textquotedblright{} systems where analytical expressions of the
system dynamics are either not available or not in a form that allows
for easy linearization~\cite{VanDerMerweWanJulier2004}.

\subsection{Distributed Hungarian-Based Algorithm}

We adopt the Distributed Hungarian-Based Algorithm (DHBA)~\cite{Ismail2017}
to implement the task-allocation function in the proposed framework.
DHBA uses an implicit coordination approach for each individual robot
to produce task-allocation results. The core the DHBA is the Hungarian
algorithm and we adopt one of its implementations as presented in
Algorithm~\ref{Hungarian Algorithm}, in which given the input as
an $m\times m$ cost matrix, $C$, a bipartite graph is constructed
based on which assignments are admissible. The vectors containing
task labels and agent labels are denoted by $U$ and $V$, both of
which are $1\times m$ vectors with elements $u_{j}$ and $v_{i}$,
respectively, where $i,j\in\{1,2,,\cdots,m\}$. The edge between $u_{j}$
and $v_{i}$ is considered admissible if $v_{i}+u_{j}=c_{ij}$. In
the bipartite graph, there are either matched admissible edges or
unmatched admissible edges (line 7). An edge is considered matched
if an agent has chosen a particular admissible task (line 12). More
details can be found in~\cite{LindsaySun2020}

\begin{algorithm}[h]
\caption{Hungarian Algorithm~\cite{papadimitriou1998combinatorial}}
\label{Hungarian Algorithm} \begin{algorithmic}[1] \Procedure{Hungarian}{$C$}\Comment{C
is $m\times m$ cost matrix of non-negative integers} \State \textbf{for
all} \texttt{$v_{i}$} $\in$ $V$ \texttt{, $v_{i}$= 0} \State
\textbf{for all} \texttt{$u_{j}$} $\in$ $U$ \texttt{, $u_{j}$=
min($C(:,j)$)} \Comment{$U$ and $V$ are task and agent label
vectors of all the tasks and agents} \While{Not Full Match} \State
\textbf{for all} \texttt{$C_{ij}$} \State \indent\textbf{if}\texttt{($v_{i}$+$u_{j}$
= $C_{ij}$):} \State \indent\indent \textit{edge is admissible}
\State \indent\textbf{end if} \State \textbf{end for} \State \textbf{for
all} \texttt{agents $i$ $\in$ $m$, tasks $j$ $\in$ $m$} \Comment{matching
phase begins} \State \indent\textbf{if}\texttt{(edge = admis. $and$
unm.):} \State \indent\indent \textbf{match} \texttt{agent $i$
and task $j$} \State \indent\textbf{end if} \State \textbf{end
for} \State \textbf{Augment}\textit{(Shortest Path)} \State \textbf{if}\texttt{(any
agent is unmatched):} \State \indent \textit{unmatched agent gets
marked} \State \indent \textit{connected nodes get marked} \State
\textbf{end if} \State \textbf{for all} \texttt{$j\in J$} \State
\indent\texttt{$slack_{j}$ = min($C_{marked,j}$)-($u_{j}$+$v_{marked}$)}
\State \textbf{end for} \State \texttt{$\delta$ = $min(slack_{j})/2$}
\State \textbf{for all} \texttt{$i\in I$ and $j\in J$} \State
\indent \textbf{if}\texttt{(agent $i$ is marked):} \State \indent\indent
\texttt{$v_{i}$=$v_{i}$+$\delta$} \State \indent \textbf{otherwise}
\State \indent\indent\texttt{$v_{i}$=$v_{i}$-$\delta$} \State
\indent\textbf{end if} \State \indent\textbf{if}\texttt{(task $j$
is marked):} \State \indent\indent \texttt{$u_{j}$=$u_{j}$-$\delta$}
\State \indent \textbf{otherwise} \State \indent\indent\texttt{$u_{j}$=$u_{j}$+$\delta$}
\State \indent\textbf{end if} \State \textbf{end for} \EndWhile
\State \textbf{return} \texttt{Match} \EndProcedure \end{algorithmic}
\end{algorithm}

\section{Methodology\label{sec:Methodology}}

In this section, we propose a framework for stochastic task allocation
that explicitly accounts for the uncertainty in situational awareness
in the assignment process. In this work, we assume that the numbers
of agents and tasks can be the same, while the proposed framework
has great potential to be applied to other cases.

The task allocation problem given situational awareness $\mathbb{S}$
is formally defined by $\left\langle f_{TG},\mathcal{T},\mathbf{P}_{\mathcal{T}},f_{\mathcal{C}},\mathcal{C},f_{TA},\Gamma,\mathbf{P}_{\Gamma}\right\rangle $
, where:
\begin{itemize}
\item $f_{TG}$ is an algorithm/function that generates task set $\mathcal{T}$
and corresponding covariance matrices set $\mathbf{P}_{\mathcal{T}}$.
\item $\mathcal{T}=\left\{ \tau_{1},\tau_{2},\cdots,\tau_{m}\right\} $
is a set of $m$ random variables, in the format of column vectors,
representing task states generated by task generator $f_{TG}$ given
$\left(\Xi,\mathbf{P}_{\Xi}\right)$.
\item $\mathbf{P}_{\mathcal{T}}=\left\{ \mathbf{P}_{\tau_{1}},\mathbf{P}_{\tau_{2}},\cdots,\mathbf{P}_{\tau_{m}}\right\} $
is a set of covariance matrices of set $\mathcal{T}$.
\item $f_{\mathcal{C}}$ is a function that generates cost matrix $\mathcal{C}$
and corresponding covariance matrices set $\mathbf{P}_{\mathcal{C}}$.
\item $\mathcal{C}=\left[c_{ij}\right]\in\mathbb{R}^{m\times m}$ is the
cost matrix, where $c_{ij}$ denotes the cost that robot $a_{i}$
pays to execute task $\tau_{j}$.
\item $\mathbf{P}_{\mathcal{C}}\in\mathbb{R}^{m^{n}\times m^{n}}$ is the
covariance matrix associated with $\mathcal{C}$.
\item $f_{TA}$ is an algorithm/function that performs task allocation.
\item $\Gamma=\left[\gamma_{ij}\right]\in\mathbb{R}^{m\times m}$ is the
resulting matrix generated by applying $f_{TA}$ onto $\mathcal{C}$
.
\item $\mathbf{P}_{\Gamma}\in\mathbb{R}^{m^{n}\times m^{n}}$ is the covariance
matrix associated with $\Gamma$.
\end{itemize}
\begin{figure}
\begin{center}\includegraphics[width=0.5\textwidth]{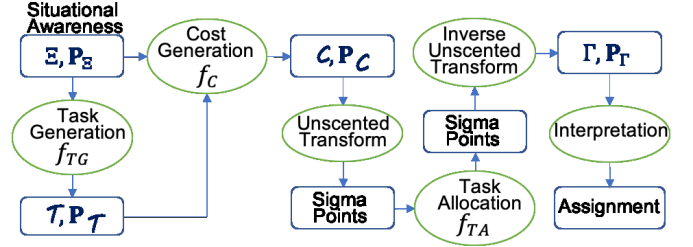}

\end{center}\caption{\label{fig:flowchart}A framework for stochastic task allocation.}

\end{figure}

The proposed framework includes the following steps (see Fig.~\ref{fig:flowchart}):
\begin{enumerate}
\item Extract the information from situational awareness ($\Xi,\mathbf{P}_{\Xi}$)
for task allocation and generate tasks in a stochastic form ($\mathcal{T},\mathbf{P}_{\mathcal{T}}$)
using function $f_{TG}$; 
\item Generate cost information ($\mathcal{C},\mathbf{P}_{\mathcal{C}}$)
using function $f_{\mathcal{C}}$; 
\item Generate sigma points using Unscented Transform described by Equations~\eqref{eq:sigma_f}
and~\eqref{eq:sigma_h}; 
\item Pass the sigma points through the task allocation process ($f_{TA}$)
and calculate the mean and covariance of the result ($\Gamma_{s},\mathbf{P}_{\Gamma_{s}}$); 
\item Generate an executable assignment $\Gamma_{f}$ by interpreting the
stochastic form of the result ($\Gamma_{s},\mathbf{P}_{\Gamma_{s}}$).
\end{enumerate}

\subsection{Interpretation Policy\label{subsec:Interpretation-Policy}}

Traditional task-allocation algorithms generate a binary assignment
matrix, whereas a saliently novel contribution of the proposed stochastic
task allocation framework is the evaluation of the confidence, $\mathbf{P}_{\Gamma_{s}}$,
on the result, $\Gamma_{s}$. Each resulting assignment via sigma
points, $\Gamma_{s,i}$, $i=0,\cdots,2L$, is a binary matrix, whereas
the result after inverse Unscented Transform, $\Gamma_{s}=\sum_{i=0}^{2L}\omega_{i}^{m}\Gamma_{s,i}$,
would be a non-binary matrix, thereby a policy will be needed to interpret
$\Gamma_{s}$ as an executable result. 

We propose a heuristic interpretation policy using the information
theory~\cite{Cox1966} to generate a binary assignment matrix. The
proposed policy is as follows:
\begin{enumerate}
\item Take the diagonal elements in $\mathbf{P}_{\Gamma}$, $\left\{ p_{11},p_{22},\cdots,p_{m^{2},m^{2}}\right\} $,
and place them in an $m\times m$ squared uncertainty matrix as
\begin{equation}
\Sigma_{s}=\left[\begin{array}{cccc}
p_{11} & p_{m+1,m+1} & \cdots & p_{m^{2}-m+1,m^{2}-m+1}\\
p_{22} & p_{m+2,m+2} & \cdots & p_{m^{2}-m+2,m^{2}-m+2}\\
\vdots & \vdots & \ddots & \vdots\\
p_{m,m} & p_{2m,2m} & \cdots & p_{m^{2},m^{2}}
\end{array}\right]
\end{equation}
\item Define a weighted inverse assignment matrix as 
\begin{equation}
Q=\left[\begin{array}{cccc}
\frac{p_{11}}{\Gamma_{s}\left[1,1\right]} & \frac{p_{m+1,m+1}}{\Gamma_{s}\left[1,2\right]} & \cdots & \frac{p_{m^{2}-m+1,m^{2}-m+1}}{\Gamma_{s}\left[1,m\right]}\\
\frac{p_{22}}{\Gamma_{s}\left[2,1\right]} & \frac{p_{m+2,m+2}}{\Gamma_{s}\left[2,2\right]} & \cdots & \frac{p_{m^{2}-m+2,m^{2}-m+2}}{\Gamma_{s}\left[2,m\right]}\\
\vdots & \vdots & \ddots & \vdots\\
\frac{p_{m,m}}{\Gamma_{s}\left[m,1\right]} & \frac{p_{2m,2m}}{\Gamma_{s}\left[m,2\right]} & \cdots & \frac{p_{m^{2},m^{2}}}{\Gamma_{s}\left[m,m\right]}
\end{array}\right],
\end{equation}
where $\Gamma_{s}\left[i,j\right]$ denotes the $ij$th element of
matrix $\Gamma_{s}$. 
\item An executable assignment matrix is given by
\begin{equation}
\Gamma_{f}=\text{HUNGARIAN}\left(Q\right).
\end{equation}
\end{enumerate}
The heuristic of this proposed policy is that the information of each
element in the assignment matrix ($\Gamma_{s}$) is the inverse of
the element in the uncertainty matrix ($\Sigma_{s}$). Each element
in the weighted inverse assignment matrix ($Q$) is the inverse of
the weighted assignment matrix, which represents the weighted uncertainty
of the assignment. Since the Algorithm~\ref{Hungarian Algorithm}
is going to minimize the overall cost, we apply it onto $Q$ to find
the assignment with the minimized overall uncertainty.

\section{Simulation Results\label{sec:Simulation-Results}}

To validate the effectiveness of the proposed framework, MATLAB simulations
were conducted. In simulation, the locations of four tasks are specified
and assumed known with no uncertainty. Four robots are deployed to
complete the tasks. The position of each robot is corrupted by Gaussian
noise, with a specified mean and variance. 

\begin{figure}
\begin{center}\includegraphics[width=0.4\textwidth]{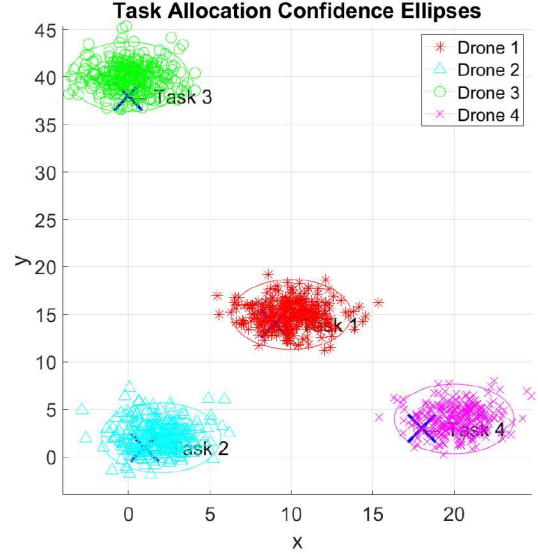}

\end{center}\caption{\label{fig:stochastic_test_case}An example simulation run showing
the locations of robots and tasks far away from each other without
overlaps of uncertainty circles. The center of a circle represents
the estimated location of a drone and the size of a circle represents
the amount of uncertainty of an estimated location.}
\end{figure}

To verify that the proposed framework generates reasonable results
in a simple case with an obvious optimal result, as shown in Figure~\ref{fig:stochastic_test_case},
the tasks (blue crosses) are placed at the following coordinates:
T1 = (9, 14), T2 = (1, 1), T3 = (0, 38), and T4 = (18, 3), . The mean
values of the $x$ and $y$ coordinates of the robots are as follows:
D1 = (10, 15), D2 = (2, 2), D3 = (0, 40), and T4 = (20, 4). As for
the covariances of each distribution, the $x$ and $y$ coordinates
are assumed to be uncorrelated, resulting in a diagonal covariance
matrix, represented by four circles of different colors. The variances
of the $x$ and $y$ coordinates are assumed to be equal, i.e., varX
= varY = 1.25. Each robot has the same covariance matrix for their
position uncertainty. It can be seen that the optimal solution for
the task allocation problem that minimizes the overall distance traveled
by each robot is to command agents 1 through 4 to travel to tasks
1 through 4, respectively. The result generated by the proposed framework
reveals the same allocation. 

\begin{figure}
\begin{center}\includegraphics[width=0.4\textwidth]{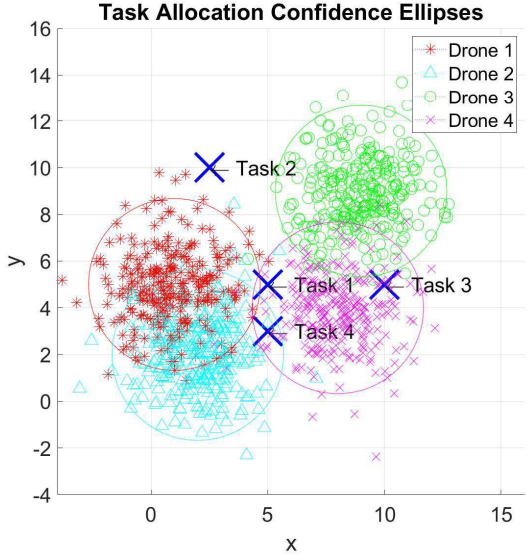}

\end{center}\caption{\label{fig:stochastic_TA}An example simulation run showing the locations
of robots and tasks are close to each other with overlaps of uncertainty
circles. The center of a circle represents the estimated location
of a drone and the size of a circle represents the amount of uncertainty
of an estimated location.}
\end{figure}

Next, we tested the proposed framework in a more complicated scenario
where positions of the robots and tasks are close to each other. Figure~\ref{fig:stochastic_TA}
shows one of many simulations in this case. The positions of the tasks
are as follows: T1 = (5, 5), T2 = (2.5, 10), T3 = (10, 5), T4 = (5,
3). The centers of the robot position distributions are as follows
respectively: D1 = (1, 5), D2 = (2, 2), D3 = (9, 9), D4 = (8, 4).
Once again, it is assumed that varX = varY = 1.25 for each robot\textquoteright s
distribution. The individual samples associated with each uncertainty
circle represent realizations of drone positions following a Gaussian
distribution. Using the center of each circle as the estimated location
of each drone, Algorithm~\ref{Hungarian Algorithm} is applied first
and obtain the assignment matrix, $\Gamma_{0}$, is given by 
\begin{equation}
\Gamma_{0}=\left(\begin{array}{cccc}
0 & 1 & 0 & 0\\
0 & 0 & 0 & 1\\
0 & 0 & 1 & 0\\
1 & 0 & 0 & 0
\end{array}\right).\label{eq:stochastic_TA}
\end{equation}
Applying the proposed stochastic task-allocation framework, the assignment
matrix $\Gamma_{s}$ and the associated uncertainty matrix $\Sigma_{s}$
are given by
\begin{align}
\Gamma_{s} & =\left(\begin{array}{cccc}
1 & -0.2 & 0 & 0.2\\
0.7 & 0 & 0 & 0.3\\
0.2 & 1.2 & -0.3 & 0\\
-0.8 & 0 & 1.3 & 0.5
\end{array}\right),\\
\Sigma_{s} & =\left(\begin{array}{cccc}
0.8 & 1.0 & 0 & 0.1\\
0.4 & 0 & 0 & 0.4\\
0.1 & 1.0 & 1.1 & 0\\
1.4 & 0 & 1.1 & 0.3
\end{array}\right).
\end{align}
The element of $\Sigma_{s}$ represents the uncertainty of the corresponding
element in $\Gamma_{s}$. Using the proposed policy in~\ref{subsec:Interpretation-Policy},
the executable assignment matrix is given by
\begin{equation}
\Gamma_{f}=\left(\begin{array}{cccc}
0 & 0 & 0 & 1\\
1 & 0 & 0 & 0\\
0 & 1 & 0 & 0\\
0 & 0 & 1 & 0
\end{array}\right).
\end{equation}

Since the actually position of each robot can be any realization in
the corresponding color zones, we conducted Monte Carlo simulations
to compare the overall cost of the assignments generated by both $\Gamma_{0}$
and $\Gamma_{f}$. The result shows that the proposed stochastic task
allocation approach generate an assignment with 30\% less overall
cost than using $\Gamma_{0}$. 

\section{Conclusion\label{sec:Conclusion}}

A novel task-allocation framework is proposed in this paper to address
the uncertainty in situational awareness for distributed autonomous
robots (DARs). The Sigma-Point samples generated by the Unscented
transform are passed through a task-allocation algorithm to generate
corresponding individual assignment matrices. The resulting overall
assignment matrix is generated by the inverse Unscented transform
and comes as a non-binary matrix. A novel interpretation policy was
also proposed to produce an executable assignment matrix. The simulation
result reveals the effectiveness of the proposed framework in Monte
Carlo simulations with 30\% reduction in the overall cost versus the
solution without considering the uncertainty in the robot states. 

\section*{Acknowledgement}

This research work was supported by the Laboratory Directed Research
and Development program at Sandia National Laboratories. Sandia National
Laboratories is a multimission laboratory managed and operated by
National Technology Engineering Solutions of Sandia, LLC, a wholly-owned
subsidiary of Honeywell International Inc., for the U.S. Department
of Energy\textquoteright s National Nuclear Security Administration
under contract DE-NA0003525. The views expressed in the article do
not necessarily represent the views of the U.S. Department of Energy
or the United States Government.

\bibliographystyle{ieeetr}
\bibliography{career_sun,sun,ICUAS20}

\end{document}